\documentclass{article} 
\usepackage{iclr2023_conference,times}

\usepackage{amsmath,amsfonts,bm}

\def\eqref#1{equation~\ref{#1}}

\def\1{\bm{1}}

\DeclareMathAlphabet{\mathsfit}{\encodingdefault}{\sfdefault}{m}{sl}
\SetMathAlphabet{\mathsfit}{bold}{\encodingdefault}{\sfdefault}{bx}{n}

\usepackage{framed, color}
\usepackage{hyperref}
\usepackage{url}
\usepackage{tabularx}
\usepackage{float}
\usepackage{array}
\usepackage{graphicx}
\usepackage{pifont}
\newcommand{\cmark}{\ding{51}}%
\newcommand{\xmark}{\ding{55}}%
\definecolor{light-gray}{gray}{0.95}
\newcommand{\code}[1]{\colorbox{light-gray}{\texttt{#1}}}

\title{AfroDigits: A Community-Driven Spoken Digit Dataset for African Languages}

\author{Chris Chinenye Emezue$^1$\thanks{Research done while interning at Hugging Face. Correspondence to chris.emezue@gmail.com.} , Sanchit Gandhi$^1$, Lewis Tunstall$^1$, Abubakar Abid$^1$, Josh Meyer$^2$, \And Quentin Lhoest$^1$, Pete Allen$^1$, Patrick von Platen$^1$, Douwe Kiela$^1$, Yacine Jernite$^1$, \And Julien Chaumond$^1$, Merve Noyan$^1$, Omar Sanseviero$^1$  \\
$^1$Hugging Face,$^2$ Coqui \\
}

\iclrfinalcopy 
\begin{document}

\maketitle


\begin{abstract}


The advancement of speech technologies has been remarkable, yet its integration with African languages remains limited due to the scarcity of African speech corpora. To address this issue, we present AfroDigits, a minimalist, community-driven dataset of spoken digits for African languages, currently covering 38 African languages. As a demonstration of the practical applications of AfroDigits, we conduct audio digit classification experiments on six African languages [Igbo (ibo), Yoruba (yor), Rundi (run), Oshiwambo (kua), Shona (sna), and Oromo (gax)] using the Wav2Vec2.0-Large and XLS-R models. Our experiments reveal a useful insight on the effect of mixing African speech corpora during finetuning. AfroDigits is the first published audio digit dataset for African languages and we believe it will, among other things, pave the way for Afro-centric speech applications such as the recognition of telephone numbers, and street numbers. We release the dataset and platform publicly at \url{https://huggingface.co/datasets/chrisjay/crowd-speech-africa} and \url{https://huggingface.co/spaces/chrisjay/afro-speech} respectively.
\end{abstract}

\section{Introduction}
\label{sec:intro}

Datasets are essential for the advancement of robust and beneficial deep neural networks in natural language processing (NLP) technologies~\cite{10.1145/3442188.3445922,masakhane}. The ImageNet~\citep{imagenet} dataset is a prime example as it revealed the power of deep neural networks in image recognition \citep{NIPS2012_c399862d, russakovsky2015imagenet}. That is to say, the more datasets there are for a given deep learning task, the better (in terms of robustness, fairness, and diversity) the model can get.


End-to-end deep learning models have pushed the state-of-the-art on speech processing tasks like automatic speech recognition (ASR) \citep{wav2vec,xlsr,radford2022robust}, and speech synthesis (TTS). However, due to data scarcity, existing speech recognition technologies do not support African languages~\citep{muhire_2020,dossou2021okwugb,afonja2021learning,sautidb}. We believe that our voice defines who we are and therefore when languages are omitted from speech technologies, the identities and cultures of the speakers are gradually obscured. The AfroDigits project was created to fill this void of African speech corpora by using a community-based participatory approach~\citep{masakhane} to build AfroDigits -- a spoken digit dataset for all African languages. This dataset has a number of potential use cases, ranging from being used to easily introduce concepts in speech processing, to real-life applications like recognition of spoken telephone digits, street house numbers, etc.

The rest of the paper is structured as follows: we motivate AfroDigits and give an overview of our data collection efforts aimed to bridge the gap for languages in speech technology. Then we detail the AfroDigits project in section \ref{sec:project}, and present the AfroDigits dataset as well as some of its useful properties in section \ref{sec:dataset}. Finally, to demonstrate a possible use-case of the dataset we perform finetuning experiments and discuss their results in section \ref{sec:exp}.

\section{Related Work}
\label{sec:related}

We focus on related efforts in building speech corpora for speech processing tasks. Some popular large-scale open-source \textit{monolingual} speech datasets, which have dominated research in speech processing, include LibriSpeech~\citep{librispeech} (as well as its variants like LibriCSS~\citep{chen2020continuous}, LibriLight~\citep{librilight} ) and TIMIT~\citep{timit}. However they do not have support for African and other non-English languages. Then came the wave of multilingual speech corpora, like Vox-Forge~\citep{voxforgeFreeSpeech}, Babel~\citep{babel}, MAILABS~\citep{caitoMAILABSSpeech} and most notably, Mozilla's Common Voice~\citep{commonvoice}, to  enable support for many more languages. However, the number of African languages supported is still meager. Out of the 2000+ African languages, only Kinyarwanda has 1000+ hours of audio on Common Voice \citep{muhire_2020}. Babel, the only project that contains a number of African languages, is 1) not open-source which limits its use to only those who can pay, and 2) has been shown to contain outdated styles of conversation that make it necessary to supplement models trained on it with datasets representing modern styles of communication for African languages~\citep{dossou2021okwugb}.

Over the years many efforts have emerged, specially to fill the void of African speech corpora~\citep{47393,dossou2021okwugb,digitalumugandaDataDigital,bibletts,sautidb,oyewusi2022tcnspeech,babirye2022building}. One notable property of some of these projects has been their use of a community-based, participatory approach to data collection. One advantage of community-based data collection is that, while being cost effective, it has been shown to ensure sustainability and scalability~\citep{adelani-etal-2022-thousand,emezue2020lanfrica,workshop2022bloom,joshi-etal-2020-state,10.1145/3442188.3445922,DVN/RXBNCZ_2022,DVN/YB9FWK_2022}. These works have mostly focused on text-speech corpora and not digits, which is what sets our work apart.

The FSDD dataset \citep{zohar_jackson_2018_1342401} which is most similar to ours in terms of the proposed use case, is English-based. Through AfroDigits, we contribute to the existing community-based efforts to build more African speech corpora, with a focus on digits.

\section{AfroDigits}
In this section, we expound on AfroDigits. This section begins with a description of the project, the interface and data curation process, and ends with an outline of the AfroDigits dataset.

\subsection{The Project}
\label{sec:project}
The dataset presented in this paper was a result of the AfroDigits project. The AfroDigits project is meant to be a life-long community-driven tool for audio digit data collection. Our motivation for choosing the domain of spoken digits lies in our desire to create an Afro-centric minimalist dataset which can easily be used for speech processing tasks (e.g for making tutorials, introducing concepts or new models, training and evaluating a model), similar to the way MNIST~\citep{mnist} is for the field of computer vision. The FSDD dataset \citep{zohar_jackson_2018_1342401} which is most similar to ours in terms of the proposed use case, is English-based. The idea is that AfroDigits will 1) inspire African researchers to learn speech processing while working on their native languages, and 2) improve the discoverability of African languages to non-African researchers and practitioners. 
Another advantage of the digits domain is that while it's harder to find sentences (especially for African languages~\citep{nllb,adebara2022afrolid}), numerical digits are universal, which makes recording straightforward.

\begin{figure}[h]
  \centering
  \centerline{\includegraphics[width=\textwidth]{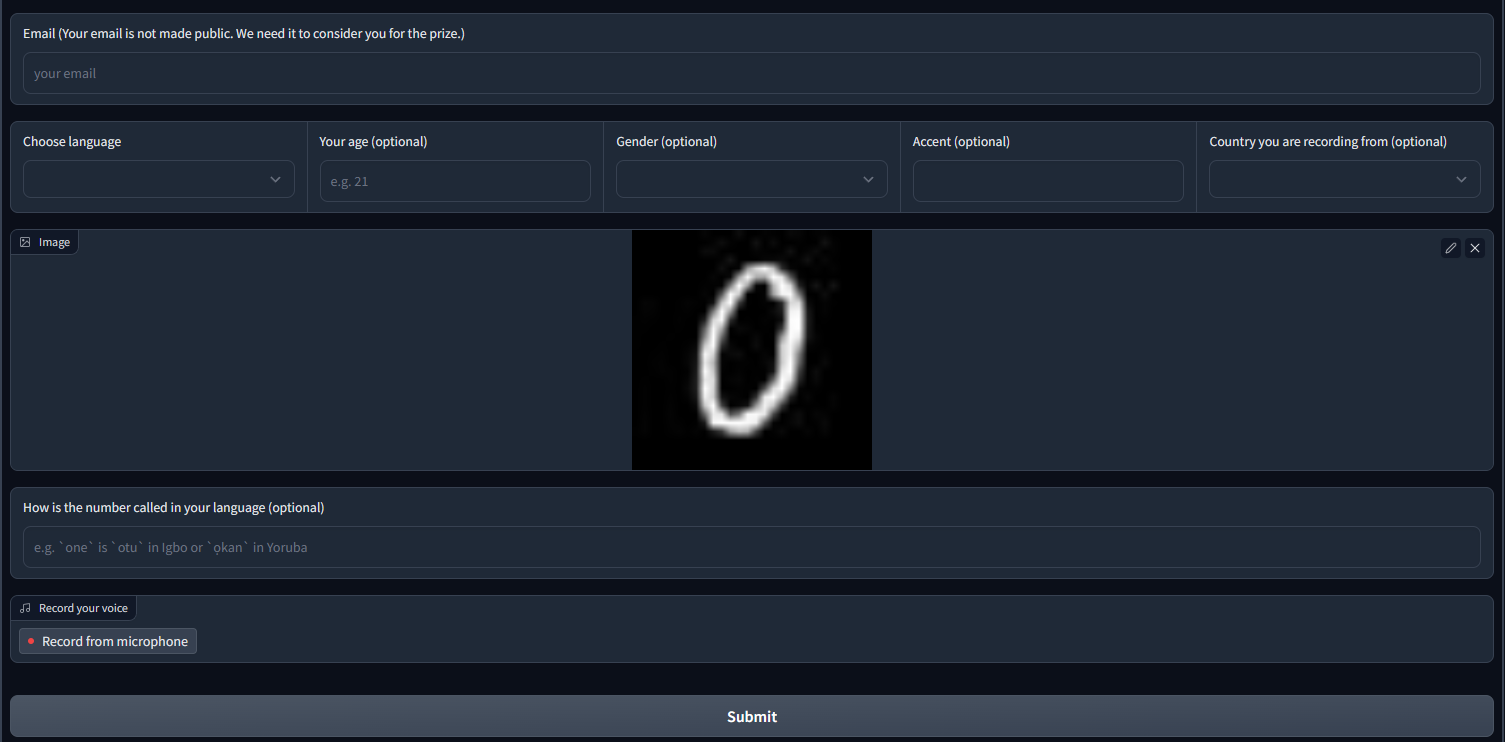}}
\caption{The AfroDigits recording platform. The participant is shown an image of the number, and recites it, while recording. The platform requires no log-in or sign-up making it very easy to use.}
\label{fig:recording-platform}
\end{figure}

\begin{figure*}[h]
  \centering
  \centerline{\includegraphics[width=\textwidth]{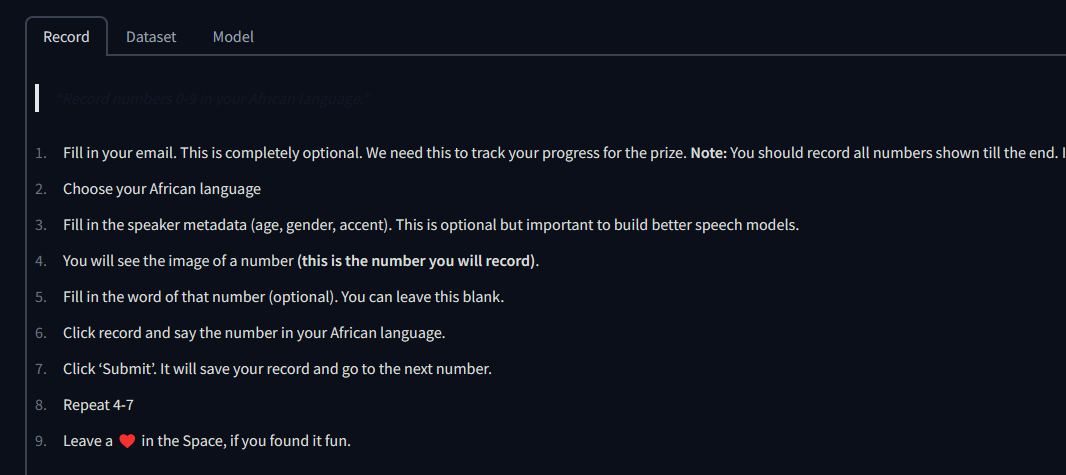}}
\caption{The set of instructions which the participant sees on entering the platform.}
\label{fig:recording-instructions}
\end{figure*}

Our first challenge was the platform to use for recording the digits. We wanted a platform that required no technical expertise to use, since we were also targeting rural communities. To this end, we created the African Digits Recording platform on HuggingFace Spaces\footnote{link to the Space will be revealed in the camera ready version}. Figure \ref{fig:recording-platform} shows the recording platform. To make the recording entertaining, and inspired by the MNIST handwritten numbers dataset, we randomly displayed images of the numbers 0-9 for the participants to record. At the end of each recording session, when they had recorded numbers 0 through 9, the participants were shown a congratulatory image GIF and encouraged to do another round. Furthermore, we wrote down very simple instructions, displayed in Figure \ref{fig:recording-instructions}, at the top of the platform so that participants would quickly understand the task.

To encourage participation during the launch of the platform, we created the African Digits Recording Sprint which lasted for one month. Through widespread advertisement, especially within communities, such as \href{www.masakhane.io}{Masakhane}, with native speakers of African languages, we ensured active participation during the sprint. We further included prizes that were given to the top ten recording contributors. In order to obtain additional meaningful metadata besides the audio, we included optional fields for users to indicate their age, gender, accent, and country of residence. Additionally, we did not require the name, address, or any other personal information of the participants, following standard practice in audio data collection~\citep{commonvoice}.

\subsection{The Dataset}
\label{sec:dataset}
Table \ref{tab:languages-all} shows the current statistics of AfroDigits, which currently has 2,185 audio samples covering 38 African languages. AfroDigits is freely available for download. Following existing data governance principles~\citep{DBLP:journals/corr/abs-1903-12262,10.1145/3531146.3534637}, the dataset is gated, meaning that one needs to provide details like name, email address and affiliation before getting access to the dataset. The whole dataset is housed in a \code{data} directory, which consists of sub-directories, each of which is named with randomly generated audio ids and contains an \code{audio.wav} file and a \code{metadata.jsonl} file where the metadata (audio id, language name, language id, digit, text of the digit, audio frequency, age, gender, and country of residence of the participant) for the specific audio file can be retrieved. All audios are mono-channel with a sampling rate of 48kHz. From Table \ref{tab:languages-all}, we see that Oshiwambo (kua) language has the highest number of recordings contributed (1,721). Using the HuggingFace Datasets\footnote{\url{https://huggingface.co/docs/datasets/index}} and Transformers\footnote{\url{https://huggingface.co/docs/transformers/index}}~\citep{wolf2019huggingfaces} libraries, one can integrate the dataset directly into their training pipeline.

\begin{table}[H]
  \caption{Current data statistics of AfroDigits. The table is sorted by the language ISO-639-3 code in alphabetical order.}
\begin{center}

 \resizebox{0.5\textwidth}{!}{%

\begin{tabular}{cccc}
 \hline
   Language & Code & \# Clips & Duration (seconds) \\
 \hline
aasáx&aas&1&2.22\\
abua&abn&10&17.1\\
abon&abo&1&2.34\\
adamorobe sign language&ads&1&19.2\\
arabic, tunisian spoken&aeb&10&23.94\\
afrikaans&afr&11&25.02\\
qimant&ahg&2&5.27\\
amharic&amh&10&25.26\\
arabic, sudanese spoken&apd&5&12.96\\
arabic, moroccan spoken&ary&10&20.4\\
arabic, egyptian spoken&arz&1&2.64\\
bambara&bam&1&2.88\\
basaa&bas&10&33.6\\
andaandi&dgl&2&5.04\\
ezaa&eza&11&76.38\\
fon&fon&3&9.6\\
oromo, borana-arsi-guji&gax&40&83.1\\
hausa&hau&1&1.74\\
igbo&ibo&138&355.95\\
kinyarwanda&kin&21&84.96\\
oshiwambo&kua&1721&3376.34\\
dholuo&luo&1&1.92\\
luwo&lwo&10&47.34\\
massalat&mdg&1&10.8\\
ndebele&nde&12&51.3\\
ndonga&ndo&1&1.38\\
!xóõ&nmn&1&2.58\\
rundi&run&35&142.79\\
shona&sna&30&70.89\\
somali&som&12&11.54\\
swahili&swa&11&11.28\\
turkana&tuv&1&1.8\\
tswapong&two&10&9.81\\
makhuwa&vmw&10&40.38\\
wolof&wol&10&9.81\\
maay&ymm&1&3.42\\
yoruba&yor&28&27.48\\
zulu&zul&1&2.68\\

\end{tabular}
  }

  \label{tab:languages-all}
\end{center}
\end{table}


\section{Experimental Setting}
\label{sec:exp}

To demonstrate the use-case of the AfroDigits dataset, we run finetuning experiments using pretrained speech models. In this section we discuss the focus languages for our experiments, as well as the models utilized.

\subsection{Focus Languages}

For our experiments, we focused on the six African languages with the most significant number of audio samples in AfroDigits -- Igbo (ibo), Yoruba (yor), Rundi (run), Oshiwambo (kua), Shona (sna), and Oromo (gax).
The distribution of the digits and gender for each language is shown in Figures~\ref{fig:ibo-digits} - \ref{fig:run-gender} in the Appendix section. Figure \ref{fig:gender} shows the gender distribution across the number of recorded clips for the focus languages. We see representation of both male and female voices in yor, ibo and kua. Together with the pie chart on the right, there is a comparably similar representation of male and female voices in all our audio samples. All this is in line with previous work~\citep{commonvoice,10.1145/3442188.3445922,masakhaner} opining that community-based crowd-sourcing fosters a wide representation of participants and improves diversity in data collection, thereby making the dataset more representative and less biased to a particular gender, race or location.

\begin{figure*}[h]
  \begin{center}
  \includegraphics[width=0.5\textwidth]
  {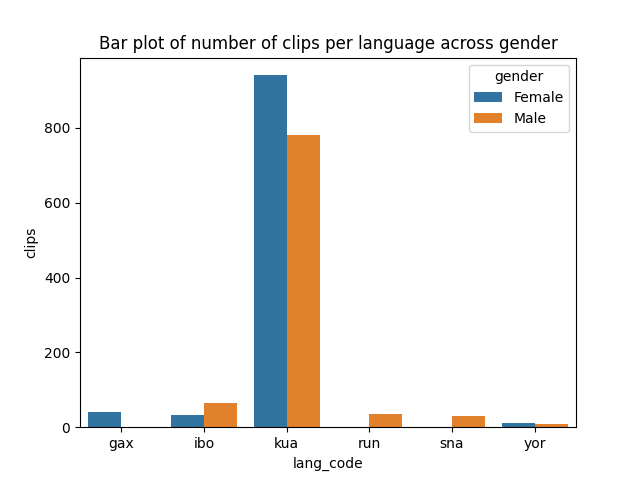}
    \includegraphics[width=0.4\textwidth]{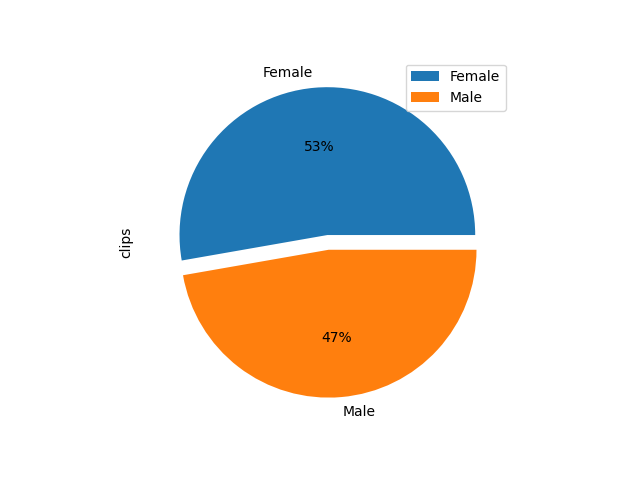}
\caption{Left: Barplot of the number of clips for each of our focus languages segmented by the gender. Right: Gender distribution of audio samples for our focus languages}
\label{fig:gender}
\end{center}
\end{figure*}

Table \ref{tab:languages} shows information about each of our focus languages. XLS-R pretraining contained audio data from three of the six focus languages (namely yor,run,and sna) while Wav2Vec2-Large has none. The African audio data used in pretraining XLS-R came mostly from Babel~\cite{babel}, which is not a free dataset. AfroDigits, being open-source and free for all, offers a major  contribution to creating open-source and free speech data for African languages.

\begin{table}[H]
\caption{Information about the languages used in our experiments, and their number of training and evaluation clips. For each pretrained model, we denote (\cmark) if the model was pretrained on audio data from that language and (\xmark) otherwise. \code{mixed} refers to the setting where we mix all the audio data from all languages.}
\begin{center}

 \resizebox{0.6\textwidth}{!}{%

\begin{tabular}{cccc}
 \hline
   Code & \#Train / \#Eval & XLS-R & Wav2Vec2-Large \\
 \hline
ibo&96 / 42&\xmark&\xmark  \\
yor&19 / 9&\cmark&\xmark  \\
run&24 / 11&\cmark&\xmark \\
kua&1204 / 517&\xmark&\xmark \\
sna&21 / 9&\cmark&\xmark \\
gax&28 / 12&\xmark&\xmark \\
\code{mixed}& 1392 / 600 & - & - \\
\hline

\end{tabular}
 }

  \label{tab:languages}
\end{center}
\end{table}

\subsection{Models}

Pretrained speech models are powerful large neural network-based models that have been trained on a gigantic speech corpora. They are trained to learn and capture meaningful abstract features from speech~\citep{schneider2019wav2vec, hsu2021robust,radford2022robust}. The knowledge learned can then be transferred to downstream tasks \citep{pmlr-v27-bengio12a,7415532}, and they are particularly useful in low-resource settings \citep{zoph2016transfer,radford2022robust,adelani-etal-2022-thousand}. This motivates our choice of using pretrained speech models on the downstream task of spoken digit classification. For our finetuning experiments, we utilized two large pretrained speech models: Wav2Vec2.0-Large \citep{wav2vec} and XLS-R \citep{xlsr}.

\paragraph{Wav2Vec2.0-Large:}  The Wav2Vec2.0-Large model was pre-trained through a self-supervised learning of representations from raw audio data by masking spans of its discretized latent speech representations, similar to masked language modeling \citep{devlin2018bert}, and using contrastive learning, where the true latent is to be distinguished from the false ones \citep{oord2018representation,wav2vec,rivire2020unsupervised,schneider2019wav2vec}. The authors note that jointly learning discrete speech units with contextualized representations helps Wav2Vec2.0 outperform the original Wav2Vec model \citep{schneider2019wav2vec} in downstream recognition tasks. The Wav2Vec2.0 model was pretrained on an English-only LibriSpeech corpus \citep{librispeech}

\paragraph{XLS-R:}
In automatic speech recognition,  researchers \citep{rivire2020unsupervised,SCHULTZ200131,1660022,6639081,hsu2021robust} have shown that it is beneficial to finetune with models that were pretrained on multilingual audio data -- especially if the multilingual corpora contains some of the languages (or language family) of your downstream language. Motivated by this, we decided to include this model in our experiments. While the Wav2Vec2.0 model was pretrained on an English-only LibriSpeech corpus \citep{librispeech}, XLS-R --  built on the Wav2Vec2.0 backbone -- was pretrained on a combined multilingual dataset of 128 languages, including 17 African languages. Table \ref{tab:languages} shows which of our target African languages are represented in the XLS-R model.






\paragraph{XLS-R-Mix: }
Some studies have underlined the relevance of mixing training datasets in NLP, particularly for low-resource languages. In machine translation, \citet{adelani2022thousand} showed that as low as $2000$ translation sentences were sufficient to effectively finetune a large pre-trained model and obtain a significantly good performance. In the speech domain, \citet{xlsr,speechstew} demonstrated that combining audio data from several languages and domains improves transfer learning capabilities in settings where the training data is very small or noisy. Inspired by this, we set out to answer the following: since the individual train samples for each language are very small (see Table \ref{tab:languages-all}), can we have some improvement, for each language, if we finetune one model on the mix of audio samples from all the languages? For this, we finetuned XLS-R on a combined dataset from all the languages using the same hyperparameters above. The resulting model is called \textit{XLS-R-Mix}.

\subsection{Training Settings}
\paragraph{Relatively Equal Model Parameters: }
The XLS-R model consists of $315,703,690$ model parameters while the Wav2Vec2.0-Large model has $315,693,962$ -- a requirement enforced in order to ensure that neither model has an edge over the other based on their size. 

\paragraph{Handling Class Imbalance:} The AfroDigits dataset currently has very small audio samples for our focus languages, with an unequal balance of digits for each language (see Figures \ref{fig:ibo-digits} - \ref{fig:run-gender}). In order to prevent the model from overfitting on the classes with many samples, we implemented weighted sampling~\citep{Monaco2013}. With weighted sampling in the data loading process, different from the normal sampling which favors the majority classes~\citep{caswell-etal-2020-language,dunn2020mapping,fan2020beyond}, the labels are chosen with a probability inversely proportional to their size in the training set. This means that at each training step and for each language, the labels with few samples are more likely to be chosen for training the model. This is similar to studies~\citep{arivazhagan2019massively,team2022language} that have leveraged upsampling for under-represented languages during pre-training large language models in machine translation.

\paragraph{Training Setup:}
All audio samples were resampled to 16kHz for the finetuning experiments. We froze the encoders of each model and finetuned for 100 epochs. We used the Adam optimizer \cite{adam}, with a learning rate of $3e-5$ for both models. We did not do any search for optimal hyperparameters but instead used the recommended settings from the authors. We ran our finetuning experiments with five different seeds, then we took the average over the different runs as well as the standard deviation. Training for each language took less than 30 minutes with a GPU, indicating the feasibility of the AfroDigits dataset as a `Hello World' dataset for speech processing, just like MNIST is for computer vision. Finetuning the dataset on large speech models like Wav2Vec2.0-Large and XLS-R, however, needs a larger GPU resource. Where needed, we used an NVIDIA A100-SXM GPU.

\section{Results \& Discussion}

\begin{figure}[h]
  \begin{center}
  \includegraphics[width=0.6\textwidth]{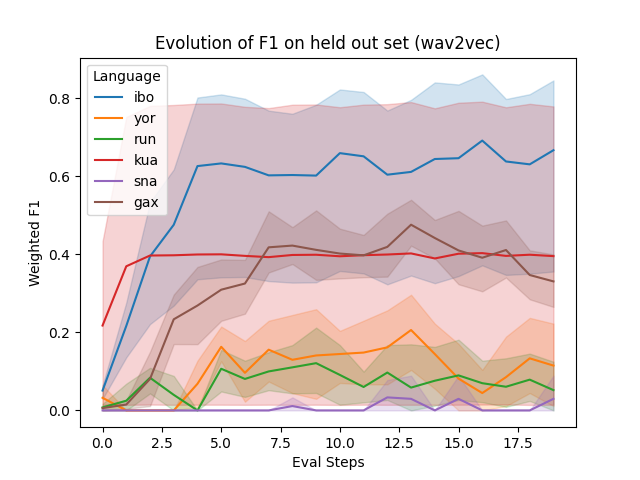}

    \includegraphics[width=0.6\textwidth]{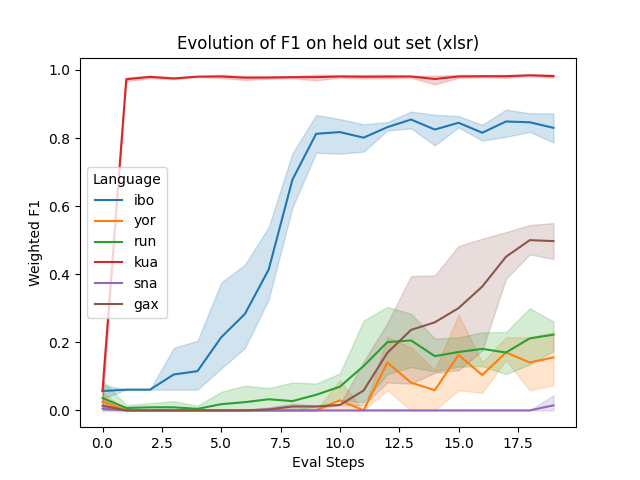}
\caption{$\uparrow$ F1 scores for each evaluation epoch during finetuning on held-out evaluation set for each of our focus African languages. Top is for Wav2Vec2.0-Large and Bottom is for XLS-R}
\label{fig:results-both}
\end{center}
\end{figure}

In Table \ref{tab:f1} we report the weighted F1 values on the held-out evaluation averaged over the 5 runs. Figure \ref{fig:results-both} shows the evolution of the finetuning performance on the held-out test set. We discuss our findings in the sub-sections below.


\begin{table}[h]
  \caption{Weighted F1 scores of each target language's evaluation set averaged over 5 runs. We see that in most cases, XLS-R performs better than Wav2Vec2-Large. XLS-R-Mix outperforms all other models in all languages.}
\centering
 \resizebox{\columnwidth}{!}{%
\begin{tabular}{ccccccc}
  \textbf{Method}& \textbf{ibo} & \textbf{yor} & \textbf{run} & \textbf{kua} & \textbf{sna}& \textbf{gax} \\
 \hline
 \\

Wav2Vec2.0-Large&$0.64 \pm 0.29$&$0.13 \pm 0.12$&$0.09 \pm 0.08$&$0.40 \pm 0.47$&$0.03 \pm 0.06$&$0.37 \pm 0.07$\\
XLS-R&$0.85 \pm 0.03$&$0.16 \pm 0.05$&$0.18 \pm 0.07$&$0.98 \pm 0.00$&$0.01 \pm 0.03$&$0.56 \pm 0.04$\\
XLS-R-Mix  &\textbf{0.86} $\pm 0.02$&\textbf{0.27} $\pm 0.07$&\textbf{0.55} $\pm 0.10$&\textbf{0.98} $\pm 0.00$&\textbf{0.65} $\pm 0.08$&\textbf{0.57} $\pm 0.01$  \\
\\
\end{tabular}
  }

  \label{tab:f1}
\end{table}

\paragraph{Bridging the gap for African Speech Datasets}
We observe, first of all, that despite being large pretrained models and finetuned for 100 epochs, the performance on some languages like yor, run, and sna is very low. This supports the claim by many research~\citep{masakhane,nllb,fan2020beyond,kreutzer2021quality,babirye2022building,oyewusi2022tcnspeech,DVN/RXBNCZ_2022,DVN/YB9FWK_2022} that there is need to build more African datasets to improve the generalization of large pretrained models to low-resource African languages, and therefore the relevance of AfroDigits.

\paragraph{Effect of having African speech data in the model pretraining: }

We observe from Table \ref{tab:f1} that the XLS-R model, which was pretrained on a larger set of African languages than Wav2Vec2-Large, performed better across all languages (except sna) than Wav2Vec2-Large which is Anglo-centric. Furthermore, Figure \ref{fig:results-both} shows the evolution of each of the model's performance (F1 metric) while it was being finetuned on each language. Each evaluation point is actually an average of the 5 rounds, with the confidence interval. Using kua as an example, we clearly see that the XLS-R model was able to quickly attain a high performance very early on in the finetuning, unlike Wav2Vec2.0-Large.

We also see that both models had difficulty with languages like sna, yor, and run: for XLS-R, we see a slight improvement for these languages only after the 50th epoch of finetuning for XLS-R, while for Wav2Vec2.0-Large, their performance oscillates between a rather low F1 score of 0.0 and 0.2. Both models perform very poorly on sna.

\paragraph{Effect of mixing audio samples during finetuning:}

XLS-R-Mix, which is XLS-R finetuned on a mix of training audio samples from the six focus languages, outperforms all the other models as shown in Table \ref{tab:f1}. More interestingly, even in sna, run, and yor, where the previous models perform very poorly, we see a significant boost in the performance of XLS-R-Mix. While the effect of a multilingual speech corpora has been shown in pretraining models \cite{xls-r,speechstew,radford2022robust}, we present a useful insight on the effect of the mixing (especially for low-resource African languages) while finetuning on spoken digit classification. 




\section{Limitations of AfroDigits}
The primary constraint observed in the initial release of the AfroDigits dataset is its small size, particularly for certain languages where only one sample is available. It is noteworthy that this project is a continuing effort and the platform used for recording voices is accessible to the general public. As such, it is anticipated that the number of recorded samples for certain languages in the dataset will expand in the future.

\section{Conclusion}
In this work, we present AfroDigits: a minimalist, community crowd-sourced dataset of recorded digits in African languages, which can scale to any African language through community effort. AfroDigits, the first African digits dataset of its kind, was created with the aim of filling the void in African speech corpora and is released as a freely accessible public dataset. We further present the current contents and statistics of AfroDigits and show spoken digits classification experiments on six African languages using the speech corpus.

\bibliography{iclr2023_conference}
\bibliographystyle{iclr2023_conference}

\appendix
\section{Appendix}

\paragraph{Analysis of gender and digits distribution in the recorded audio samples: } In the figures below, we plot their distribution, for both gender and the digits, across the Igbo, Oshiwambo and Rundi languages.
\begin{figure*}[h]
  \centering
  \centerline{\includegraphics[width=\textwidth]{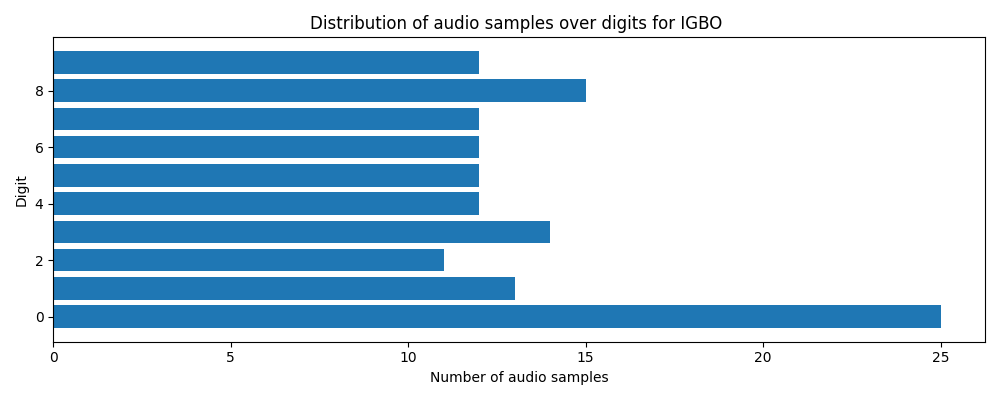}}
\caption{Distribution of recorded audio samples across the digits (ibo)}
\label{fig:ibo-digits}
\end{figure*}

\begin{figure*}[h]
  \centering
  \centerline{\includegraphics[width=\textwidth]{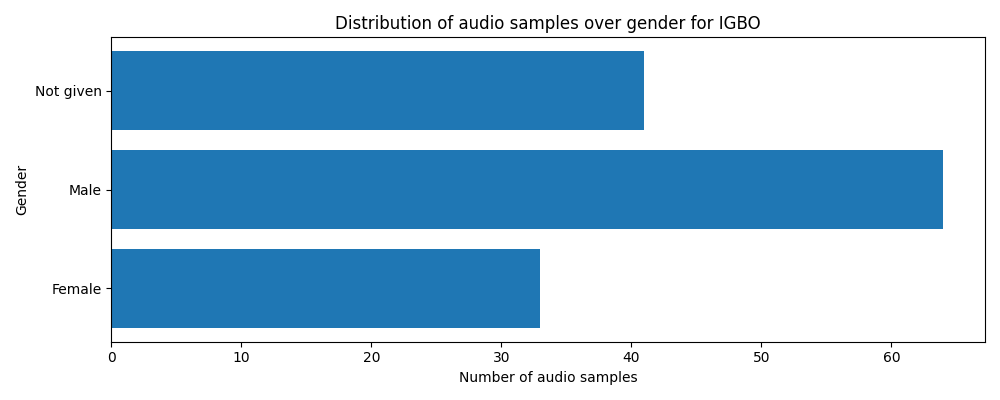}}
\caption{Distribution of recorded audio samples across the gender (ibo)}
\label{fig:ibo-gender}
\end{figure*}


\begin{figure*}[h]
  \centering
  \centerline{\includegraphics[width=\textwidth]{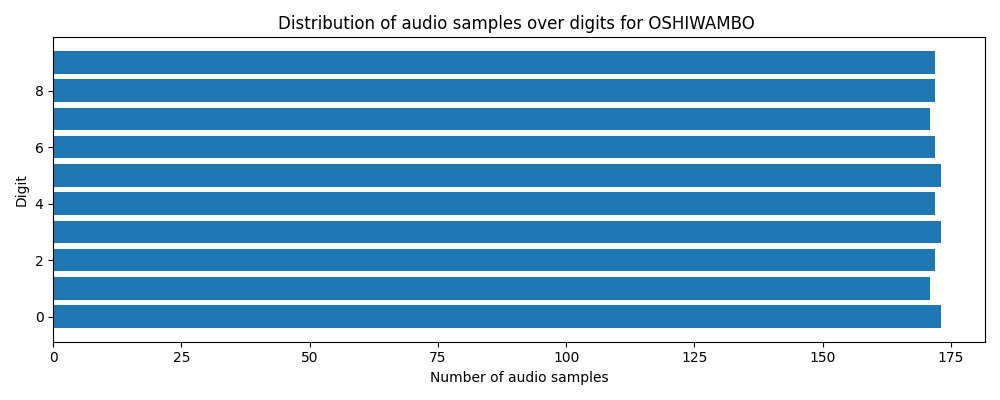}}
\caption{Distribution of recorded audio samples across the digits (kua)}
\label{fig:kua-digits}
\end{figure*}

\begin{figure*}[h]
  \centering
  \centerline{\includegraphics[width=\textwidth]{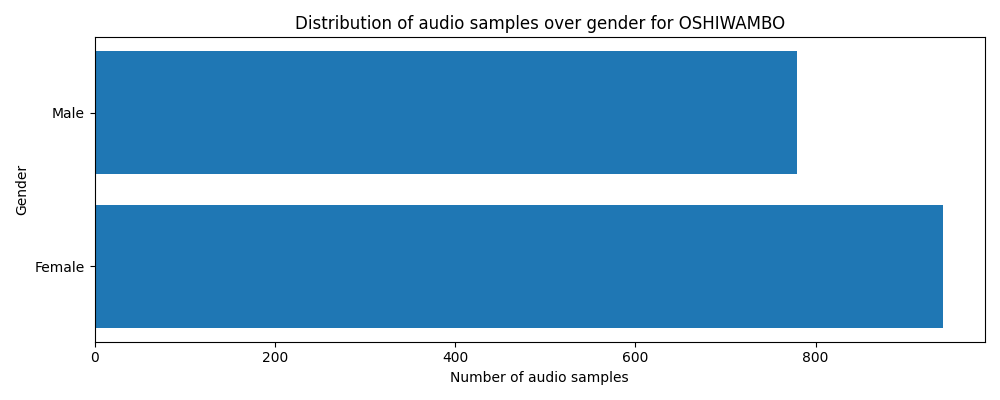}}
\caption{Distribution of recorded audio samples across the gender (kua)}
\label{fig:kua-gender}
\end{figure*}



\begin{figure*}[h]
  \centering
  \centerline{\includegraphics[width=\textwidth]{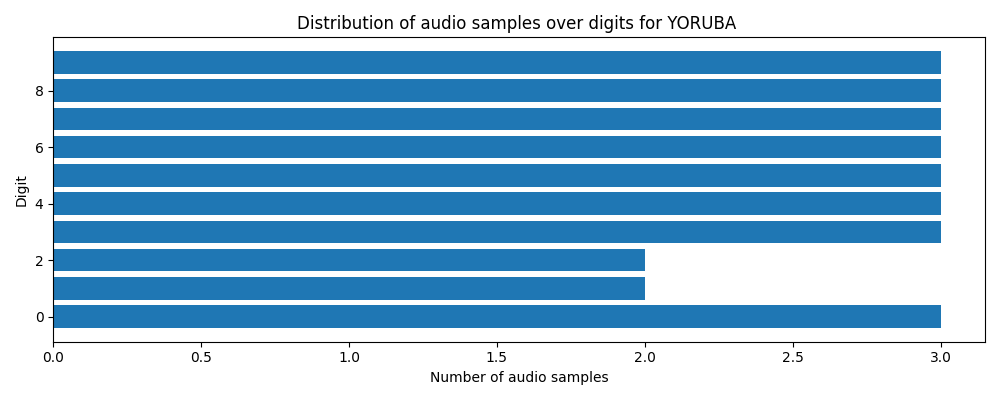}}
\caption{Distribution of recorded audio samples across the digits (yor)}
\label{fig:yor-digits}
\end{figure*}

\begin{figure*}[h]
  \centering
  \centerline{\includegraphics[width=\textwidth]{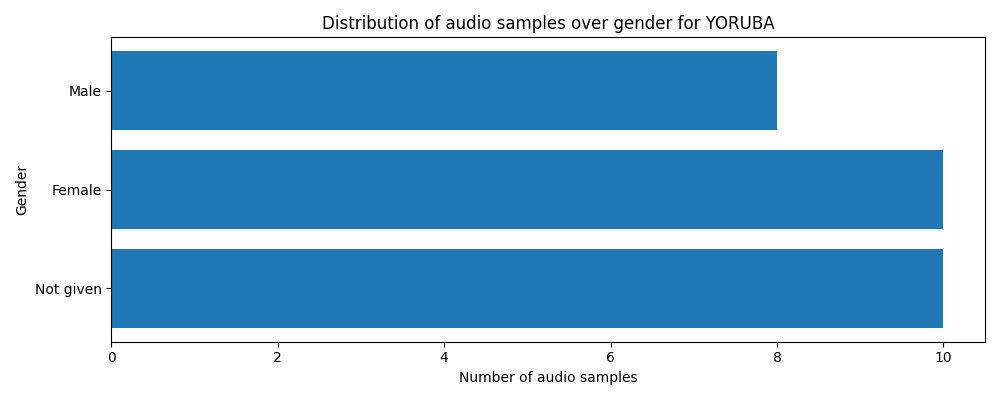}}
\caption{Distribution of recorded audio samples across the gender (yor)}
\label{fig:yor-gender}
\end{figure*}


\begin{figure*}[h]
  \centering
  \centerline{\includegraphics[width=\textwidth]{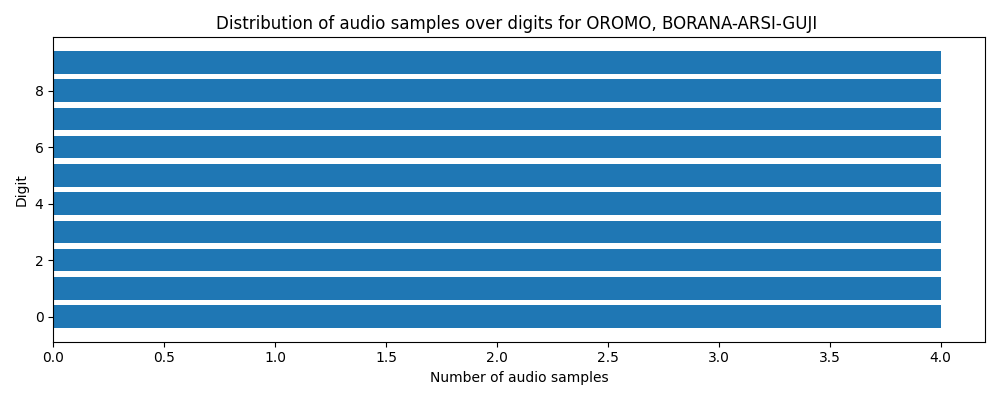}}
\caption{Distribution of recorded audio samples across the digits (gax)}
\label{fig:gax-digits}
\end{figure*}

\begin{figure*}[h]
  \centering
  \centerline{\includegraphics[width=\textwidth]{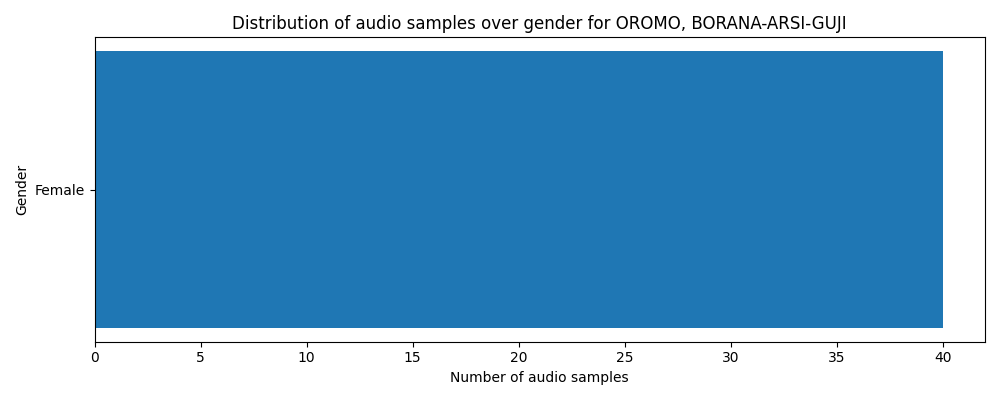}}
\caption{Distribution of recorded audio samples across the gender (gax)}
\label{fig:gax-gender}
\end{figure*}


\begin{figure*}[h]
  \centering
  \centerline{\includegraphics[width=\textwidth]{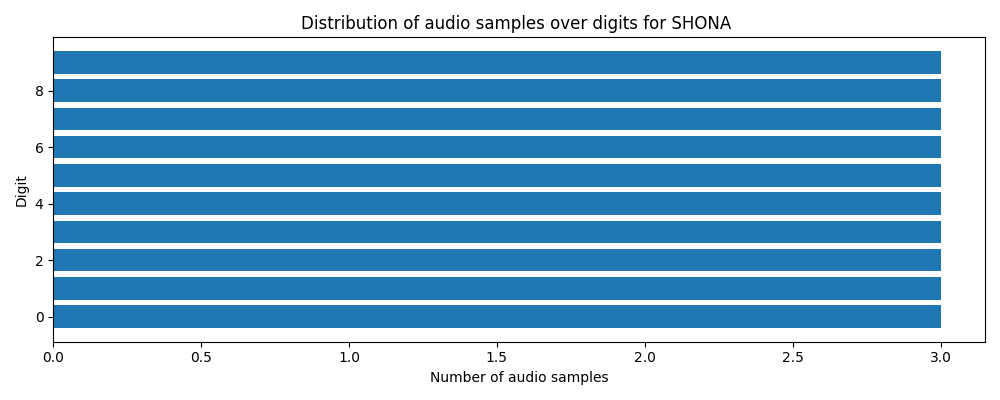}}
\caption{Distribution of recorded audio samples across the digits (sna)}
\label{fig:sna-digits}
\end{figure*}

\begin{figure*}[h]
  \centering
  \centerline{\includegraphics[width=\textwidth]{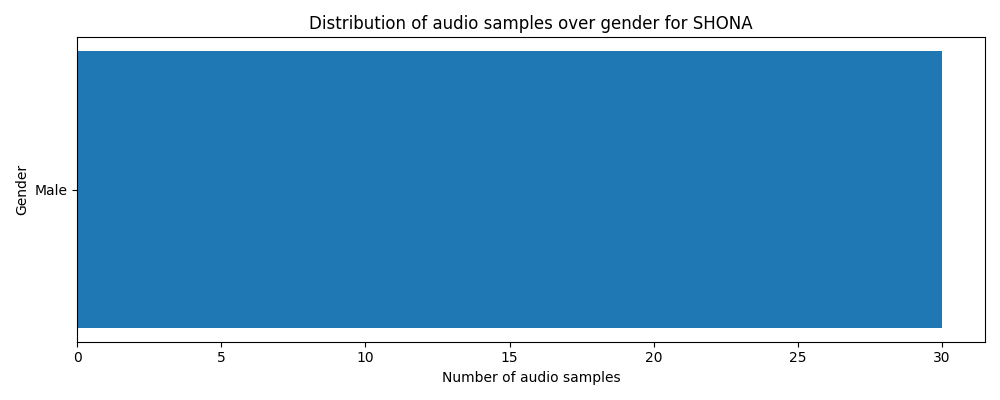}}
\caption{Distribution of recorded audio samples across the gender (sna)}
\label{fig:sna-gender}
\end{figure*}


\begin{figure*}[h]
  \centering
  \centerline{\includegraphics[width=\textwidth]{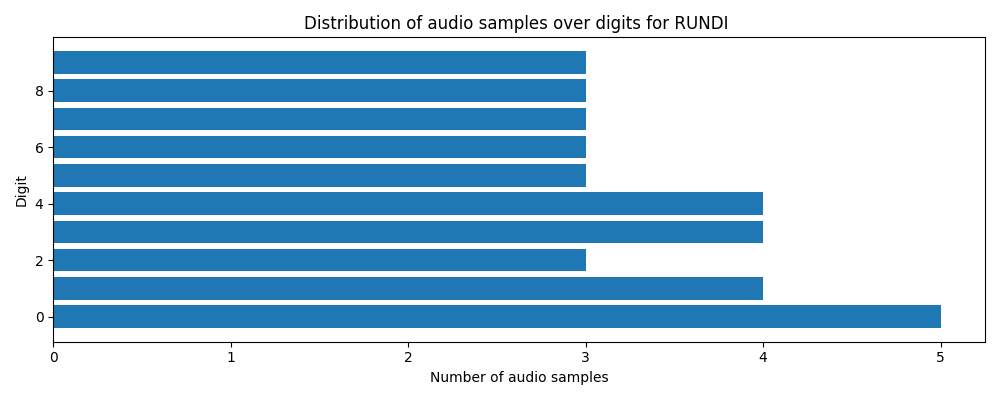}}
\caption{Distribution of recorded audio samples across the digits (run)}
\label{fig:run-digits}
\end{figure*}

\begin{figure*}[h]
  \centering
  \centerline{\includegraphics[width=\textwidth]{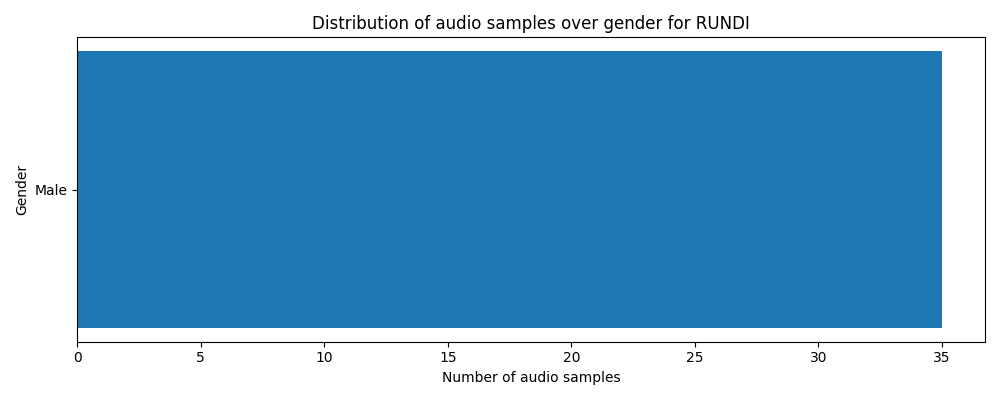}}
\caption{Distribution of recorded audio samples across the gender (run)}
\label{fig:run-gender}
\end{figure*}

\end{document}